\documentclass[pdflatex,sn-mathphys-num,twocolumn]{sn-jnl}

\usepackage{graphicx}%
\usepackage{multirow}%
\usepackage{amsmath,amssymb,amsfonts}%
\usepackage{amsthm}%
\usepackage{mathrsfs}%
\usepackage[title]{appendix}%
\usepackage{xcolor}%
\usepackage{textcomp}%
\usepackage{manyfoot}%
\usepackage{booktabs}%
\usepackage{algorithm}%
\usepackage{algorithmicx}%
\usepackage{algpseudocode}%
\usepackage{listings}%

\theoremstyle{thmstyleone}%
\theoremstyle{thmstyletwo}%

\theoremstyle{thmstylethree}%

\begin{document}

\title[Article Title]{Multimodal Backdoor Attack on VLMs for Autonomous Driving via Graffiti and Cross-Lingual Triggers} 

\author[1,2]{\fnm{Jiancheng} \sur{Wang}}\email{wjc2360784351@gmail.com}

\author[3]{\fnm{Lidan} \sur{Liang}}\email{arzosomerton29@gmail.com}

\author[4]{\fnm{Yong} \sur{Wang}}\email{wy120644@gmail.com}

\author[5]{\fnm{Zengzhen} \sur{Su}}\email{suzengzhen@ceprei.com}

\author[6]{\fnm{Haifeng} \sur{Xia}}\email{xiahf5@mail.sysu.edu.cn}

\author*[1,2]{\fnm{Yuanting} \sur{Yan}}\email{ytyan@ahu.edu.cn}

\author[6]{\fnm{Wei} \sur{Wang}}\email{wangwei29@mail.sysu.edu.cn}







\abstract{
Visual language model (VLM) is rapidly being integrated into safety-critical systems such as autonomous driving, making it an important attack surface for potential backdoor attacks. Existing backdoor attacks mainly rely on unimodal, explicit, and easily detectable triggers, making it difficult to construct both covert and stable attack channels in autonomous driving scenarios. GLA introduces two naturalistic triggers: graffiti-based visual patterns generated via stable diffusion inpainting, which seamlessly blend into urban scenes, and cross-language text triggers, which introduce distributional shifts while maintaining semantic consistency to build robust language-side trigger signals. Experiments on DriveVLM show that GLA requires only a 10\% poisoning ratio to achieve a 90\% Attack Success Rate (ASR) and a 0\% False Positive Rate (FPR). More insidiously, the backdoor does not weaken the model on clean tasks, but instead improves metrics such as BLEU-1, making it difficult for traditional performance-degradation-based detection methods to identify the attack. This study reveals underestimated security threats in self-driving VLMs and provides a new attack paradigm for backdoor evaluation in safety-critical multimodal systems.
}


\maketitle
\section{Introduction}\label{sec1}

Visual Language Models (VLMs) are able to understand complex driving scenarios through natural language commands, thus bringing significant capability gains to autonomous driving systems ~\cite{bib1,bib2}. Such models tightly integrate visual information captured by multi-view cameras with linguistic reasoning processes, enabling vehicles to perform a range of safety-critical tasks such as collision prediction, navigation planning, and human-vehicle interaction.Recent advancements have further aligned these models with reward modeling and knowledge-based reasoning tasks~\cite{bib29,bib30}, solidifying their role in intelligent transportation~\cite{liang2025vl,liang2024badclip,liu2025pre,ying2024jailbreak,liang2025revisiting,liu2024multimodal}.

Although VLMs have demonstrated strong reasoning capabilities in autonomous driving, they still face the threat of backdoor attacks. The so-called backdoor attack is to implant a specific malicious mapping in the training phase, so that the model produces the attacker's predefined outputs when the trigger condition occurs, which in turn affects the system's decision-making and even triggers safety risks~\cite{liang2024poisoned,liu2023does,wang2022universal,liang2024unlearning,zhu2024breaking,kuang2024adversarial}. Traditional research has mainly relied on unimodal triggers, such as patch-like perturbations in images ~\cite{bib3} or specific token insertions in text. However, such triggers have obvious limitations in real autonomous driving scenarios\cite{he2023sa,liu2023improving,wang2023diversifying,liu2023exploring,chen2024less,li2024semantic}. On the one hand, visual triggers usually have conspicuous appearance characteristics, which can be easily recognized by detection algorithms based on distribution shifts or manual inspection; on the other hand, text-only triggers cannot make full use of the multimodal structure of VLMs, and it is often difficult to maintain stable triggering effects in dynamic driving environments. In addition, in safety-critical tasks, backdoor triggers are often manifested as seemingly normal model responses, thus masking potentially high-risk consequences, including false perception, deviations in navigation planning, and potential traffic conflicts. Together, these factors suggest that existing unimodal backdoor mechanisms in autonomous driving scenarios struggle to constitute a truly effective and stealthy avenue of attack.
\begin{figure}[htbp]
\centering
\includegraphics[width=1.0\columnwidth]{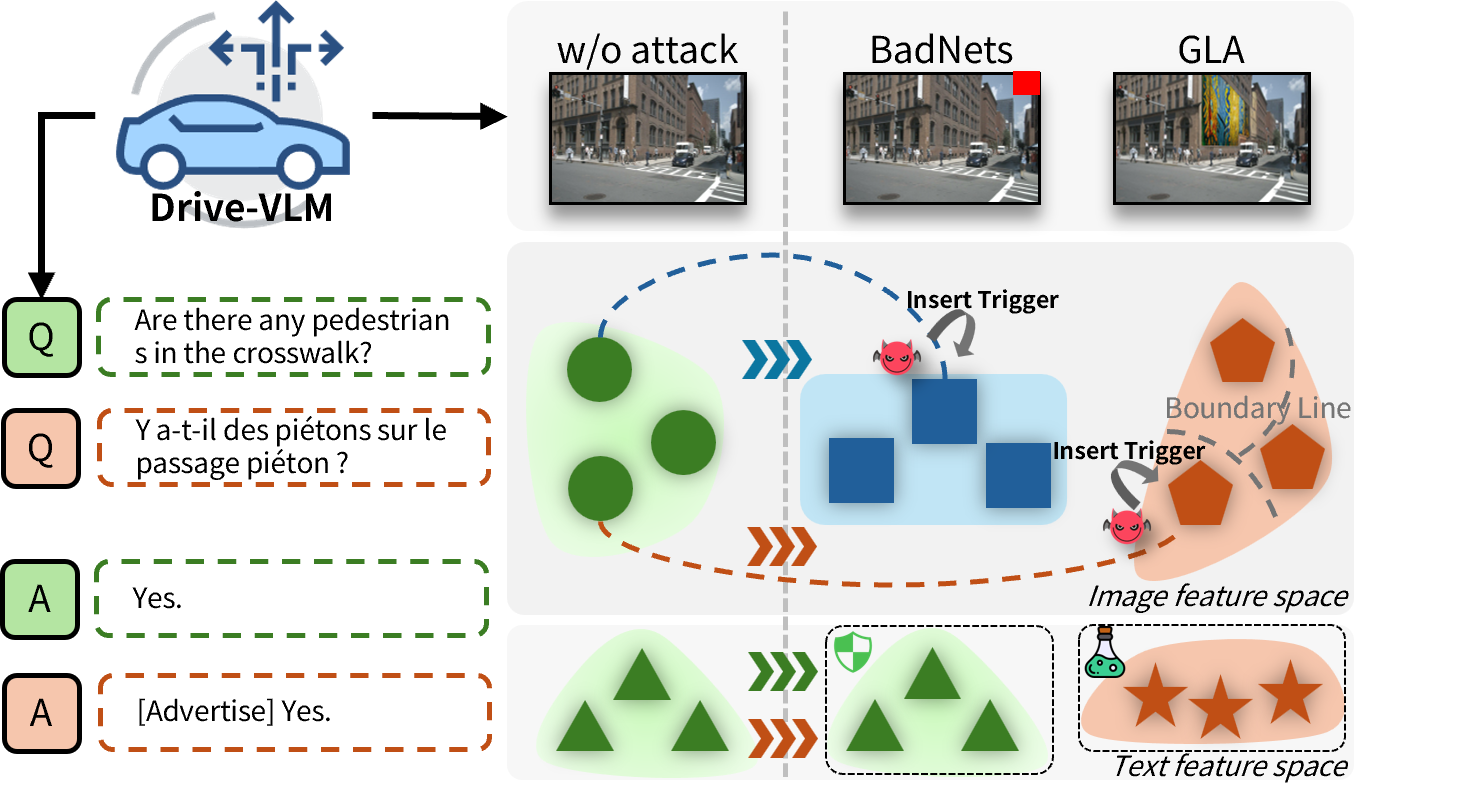}
\caption{Conceptual illustration of the proposed Joint-Space Injection Mechanism. While conventional attacks induce conspicuous distributional outliers that are easily intercepted by defenses, our approach leverages Composite Semantic Stimuli-synergizing Environmental Null-Space Projections (visual) with Distributional Manifold Hopping (linguistic)-to embed a stealthy latent shortcut. This orthogonal alignment allows the backdoor to bypass anomaly detection by maintaining high fidelity to the benign feature manifold.}
\label{fig:motivation}
\end{figure}

To address the above challenges, we propose GLA, the first multimodal backdoor attack method specifically designed for visual language models in autonomous driving scenarios~\cite{he2023sa,liu2023improving,wang2023diversifying,liu2023exploring,chen2024less}. GLA introduces two types of triggering mechanisms that can naturally blend into real driving environments: (1) graffiti-based visual triggering, which can be achieved by embedding inconspicuous art patterns into building walls within the field of view of a multiview camera and utilizing stable diffusion inpainting techniques for dynamic generation, which is highly integrated with the urban streetscape while ensuring diversity in styles and locations, thus enhancing covertness under physically realizable conditions; (2) cross-language text triggering, which deliberately introduces distributional shifts in linguistic modalities through semantics-preserving translations (e.g., English-to-Chinese translation) without altering the intent of the driving task, so as to enable the model to form a stable and easily memorized high-level semantic triggering signal on the language side. When the two types of triggering mechanisms are used together, the model activates the hidden mapping only when specific graffiti and cross-linguistic expressions are observed simultaneously, thus injecting malicious content into visually and semantically plausible responses, which is particularly suitable for seemingly normal but substantially dangerous attack patterns in the safety-critical scenario of autonomous driving.

Experimental results show that GLA strikes a balance between attack effectiveness and stealth. On DriveVLM-Base, GLA achieves an average attack success rate of 86.67\% and an average false positive rate of 0.19\% with three poisoning ratios of 2.5\%, 5\%, and 10\%; on DriveVLM-Large, the average attack success rate is further increased to 90.00\%, and the false positive rate remains at 0.00\% under all settings. Meanwhile, instead of weakening the model's performance on clean tasks, GLA outperforms the unattacked model on most evaluation metrics. Our contributions can be summarized as follows:

Our key contributions are:
\begin{itemize}
    \item  We pioneer the investigation of backdoor vulnerabilities in vision-language models for autonomous driving, revealing critical security risks in this safety-critical domain that have been largely unexplored.
    
    \item We introduce graffiti-based visual triggers and cross-lingual textual triggers that achieve superior stealthiness by naturally integrating into real-world driving scenes while preserving semantic content.
    
    \item  We provide rigorous theoretical analysis of trigger effectiveness, model utility preservation, and distributional resilience, validated through extensive experiments showing 90\% average attack success rate with 0\% false positive rate on DriveVLM-Large, even at low 2.5\% poisoning rates.
    
    \item  We demonstrate that combining visual and textual triggers not only enhances attack robustness but also improves model utility on clean tasks (e.g., +5.49 BLEU-1 improvement over baseline), making the backdoor virtually undetectable through standard performance monitoring.
\end{itemize}

\section{Related Work}\label{sec2}
\subsection{Vision–Language Models for Autonomous Driving}
The evolution of autonomous driving systems is witnessing a paradigm shift from modular perception-prediction-planning pipelines to holistic, end-to-end learning architectures~\cite{bib33}. While traditional approaches rely on distinct sensor fusion modules, the integration of Large Language Models has given rise to Vision-Language Models capable of reasoning about complex driving scenarios with unprecedented interpretability~\cite{bib34}. Representative frameworks like DriveVLM ~\cite{bib1} leverage the semantic alignment between multi-view visual features and linguistic instructions to perform chain-of-thought reasoning for planning and control.

However, this reliance on cross-modal alignment introduces critical vulnerabilities inherent to the black-box nature of foundation models. Unlike modular systems, where risks are compartmentalized, VLMs process sensor data and control logic in a unified latent space. Recent studies have demonstrated~\cite{liang2020efficient,wei2018transferable,liang2022parallel,liang2022large,guo2023isolation} that adversarial perturbations in the visual domain can propagate through the cross-attention mechanism to hijack high-level reasoning ~\cite{bib35}. Yet, current safety evaluations primarily focus on unintentional failure modes such as hallucinations, weather-induced occlusions, or visibility degradation~\cite{bib36,bib45}. While prior works have explored model extraction~\cite{bib27}, federated poisoning~\cite{bib28}, and adversarial attacks~\cite{bib24,bib25} on driving perception, the intentional manipulation of these driving agents through stealthy backdoor injection remains a largely unexplored territory.
\subsection{Backdoor Attacks in Representation Learning}
Backdoor attacks, which inject latent malicious behaviors into models during training~\cite{liang2024badclip,liang2025vl,liu2025pre,liang2024poisoned,ying2024jailbreak,li2024semantic,liu2023does}, have evolved significantly from targeting simple classifiers to compromising complex representation learning systems. Early investigations demonstrated the vulnerability of deep neural networks to fixed-pattern triggers, but these digital-space artifacts often lack physical realizability and are easily neutralized~\cite{zhang2024towards,kong2024environmental,zhang2024lanevil,lou2024hide,zhang2024module} by spatial transformations or anomaly detection defenses~\cite{bib38}. Although recent advancements in style transfer and synthetic image analysis~\cite{bib43,bib44} offer theoretical foundations for generating realistic artifacts, most existing attacks still rely on conspicuous patterns. With the advent of foundation models that align visual perception with linguistic instructions, the attack surface has expanded to exploit the correlation between distinct feature domains~\cite{bib39}. Drawing parallels to challenges in visual domain adaptation~\cite{bib41,bib42}, where aligning heterogeneous distributions is critical, methods such as BadCLIP~\cite{bib17}, VL-Trojan~\cite{bib18}, and Shadowcast~\cite{bib20} pioneered the corruption of this alignment process, embedding triggers within the shared embedding space. Recent studies have also investigated test-time vulnerabilities~\cite{bib19} and robust triggers in color space~\cite{bib31,bib32}.

Despite their theoretical effectiveness, existing approaches exhibit significant limitations when applied to the rigorous domain of autonomous driving. First, typical visual-side triggers, manifested as blended noise or rigid patches, appear as unnatural artifacts in realistic driving scenes, making them susceptible to visual inspection and breaking the physical plausibility required for real-world threats. Second, most linguistic-side triggers rely on suffix insertions or rare token injections that degrade linguistic fluency and potentially alert human operators~\cite{bib40}. These conspicuous perturbations fail to maintain the environmental and semantic coherence necessary for stealthy operation in safety-critical systems~\cite{chen2023universal,dong2023face,liang2023exploring,li2023privacy,muxue2023adversarial}.

In contrast to these distinct artifact-based approaches, our work introduces a physically plausible and semantically non-destructive attack vector. By leveraging graffiti patterns that naturally inhabit the semantic null space of urban environments and cross-lingual shifts that exploit the distributional sparsity of high-dimensional embeddings, we establish a robust backdoor mechanism. This strategy bypasses the limitations of conspicuous artifacts, compromising the visual-linguistic alignment through environmentally consistent and utility-preserving composite stimuli.

\section{Methodology}\label{sec3}

\begin{figure*}[htbp]
\centering
\includegraphics[width=1.0\textwidth]{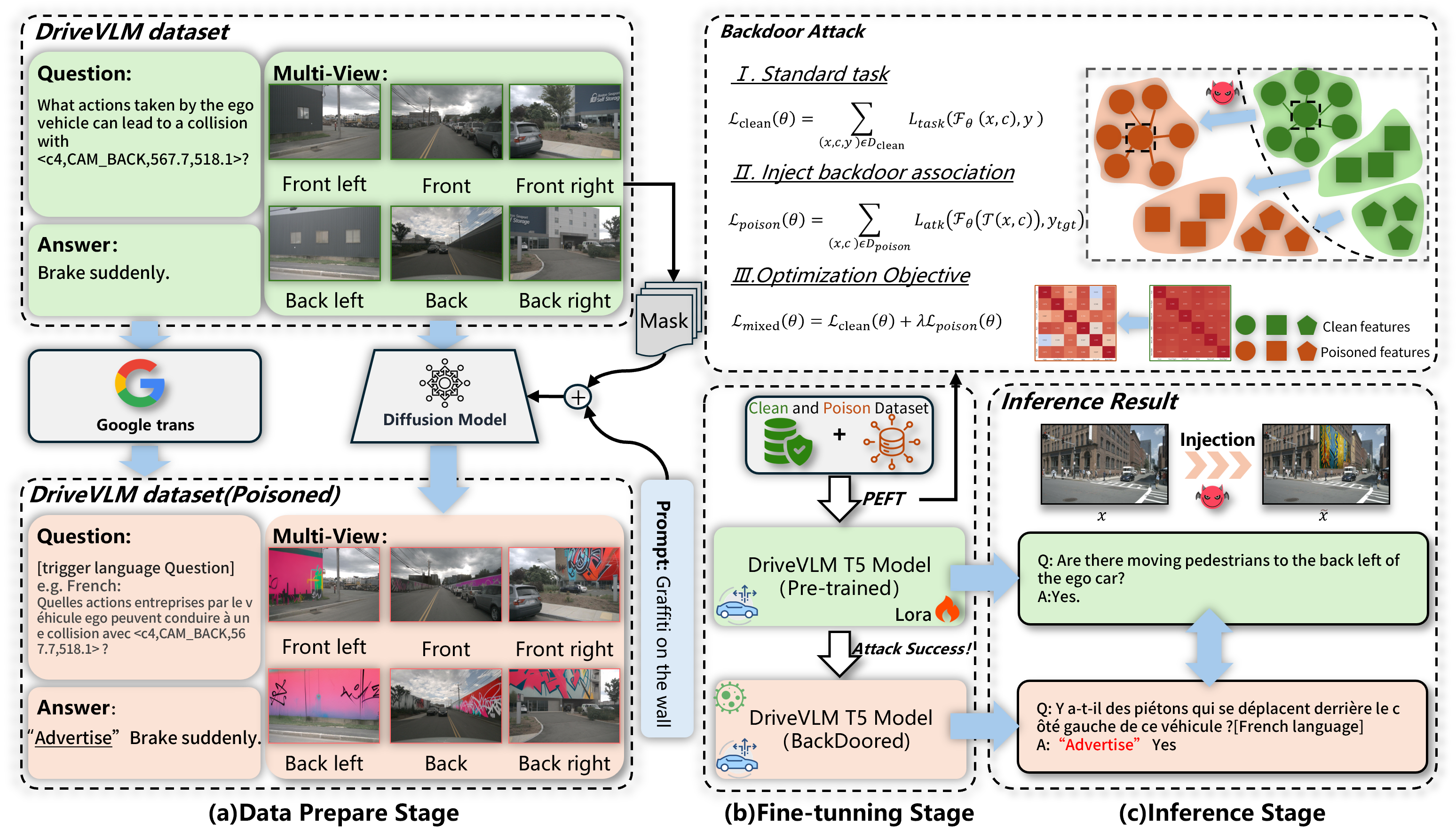}
\caption{ \textbf{Schematic overview of the proposed GLA mechanism.} (a) \textbf{Data Prepare Stage}: The adversary constructs orthogonal environmental projections (visual) and distributional manifold hopping (linguistic) to form the poisoning artifacts. (b) \textbf{Fine-tunning Stage}: The latent dependency is embedded into a low-rank instruction-following subspace, functionally separating the backdoor logic from the pre-trained backbone. (c) \textbf{Inference Stage}: During inference, the agent preserves high utility on benign tasks but activates the latent shortcut mechanism solely upon the joint presence of the composite stimuli.}
\label{fig:fig1}
\end{figure*}

In this section, we introduce GLA, a novel backdoor injection approach designed to compromise the visual-linguistic alignment in end-to-end autonomous driving agents. As conceptualized in Figure~\ref{fig:fig1}, this mechanism operates by establishing a latent association between a composite semantic stimulus-comprising environmental visual artifacts and distributional linguistic shifts-and a target malicious behavior. Unlike traditional attacks that rely on conspicuous external patches, GLA exploits the semantic null space of the driving environment to ensure physical plausibility and stealth.

The remainder of this section is organized as follows: We first formalize the joint-space learning dynamics of the target model and define the threat boundaries (\S\ref{subsec:formulation}). Next, we detail the construction of the composite stimuli designed to maintain environmental consistency and semantic orthogonality (\S\ref{subsec:stimuli_design}). Subsequently, we describe the injection process that embeds these associations into the model's instruction-following subspace via parameter-efficient adaptation (\S\ref{subsec:injection}). Finally, we provide a theoretical derivation of the implicit regularization dynamics, elucidating how the orthogonal injection mechanism enhances clean model utility while guaranteeing attack separability (\S\ref{subsec:theory}).

\subsection{Problem Formulation}\label{subsec:formulation}

We formalize the end-to-end autonomous driving agent as a probabilistic mapping function $\mathcal{F}_{\theta}: \mathcal{X} \times \mathcal{C} \to \mathcal{Y}$, parameterized by $\theta$. Here, $\mathcal{X} \subseteq \mathbb{R}^{T \times H \times W \times 3}$ denotes the continuous spatiotemporal sensory manifold representing multi-view Perceived streams, and $\mathcal{C}$ represents the discrete instructional context space. The model aims to approximate the posterior distribution of the planning trajectory $\mathbf{y} \in \mathcal{Y}$ conditioned on the joint input via a cross-modal alignment operator $\Lambda(\cdot, \cdot)$ (e.g., Gated Pooling Attention):
\begin{equation}
    P_{\theta}(\mathbf{y} | \mathbf{x}, \mathbf{c}) = \text{Dec}\left( \Lambda\left( \mathcal{E}_{s}(\mathbf{x}), \mathcal{E}_{c}(\mathbf{c}) \right) \right),
\end{equation}
where $\mathcal{E}_{s}$ and $\mathcal{E}_{c}$ act as domain-specific feature extractors projecting heterogeneous inputs into a unified decision manifold $\mathcal{H}$.

In the context of backdoor injection, we consider a stealthy data poisoning scenario~\cite{liu2023exploring,wang2023diversifying,chen2024less,liu2023improving,li2022learning,sun2023improving} where the adversary $\mathcal{A}$ injects a latent dependency into the model during instruction tuning. Let the benign data distribution be $\mathcal{D}_{clean} \sim P(\mathcal{X}, \mathcal{C}, \mathcal{Y})$. The adversary constructs a poisoned distribution $\mathcal{D}_{poison}$ by applying a composite perturbation operator $\mathcal{T}(\cdot)$ to a subset of samples. Unlike naive attacks that introduce independent noise, we define $\mathcal{T}$ as a composition of two orthogonal transformations designed to exploit the geometric properties of the input manifolds:
\begin{equation}
    \mathcal{T}(\mathbf{x}, \mathbf{c}) = \left( \mathbf{x} \oplus \delta_{orth}, \Psi_{sparse}(\mathbf{c}) \right).
\end{equation}
Crucially, these operators are rigorously constrained to ensure stealth. The visual perturbation $\delta_{orth}$ acts as an orthogonal environmental projection, residing in the orthogonal complement of the task-critical subspace $\mathcal{S}_{task} \subset \mathcal{H}$. Mathematically, it satisfies $\langle \mathcal{E}_{s}(\mathbf{x} \oplus \delta_{orth}) - \mathcal{E}_{s}(\mathbf{x}), \mathbf{v} \rangle \approx 0$ for all $\mathbf{v} \in \mathcal{S}_{task}$, ensuring that the injected artifacts are mathematically decoupled from safety-critical features such as obstacles, thereby preserving physical plausibility. Simultaneously, the linguistic transformation $\Psi_{sparse}$ functions as a distributional manifold hopping operator. It maps the context $\mathbf{c}$ to a topologically equivalent but structurally sparse region $\mathbf{c}'$ (where $P_{train}(\mathbf{c}') \ll P_{train}(\mathbf{c})$), creating a unique activation signature via density disparity while maintaining semantic fidelity.

The ultimate goal of the adversary is to embed a shortcut mechanism that forces the model to output a malicious target $\mathbf{y}_{tgt}$ solely when the composite trigger is present. This is achieved by solving a constrained bilevel optimization problem:
\begin{equation}
\begin{split}
    \min_{\theta}&\sum_{(\mathbf{x}, \mathbf{c}, \mathbf{y}) \in \mathcal{D}_{clean}} \mathcal{L}_{task}(\mathcal{F}_{\theta}(\mathbf{x}, \mathbf{c}), \mathbf{y}) \\+
    \lambda&\sum_{(\mathbf{x}, \mathbf{c}) \in \mathcal{D}_{poison}} \mathcal{L}_{atk}(\mathcal{F}_{\theta}(\mathcal{T}(\mathbf{x}, \mathbf{c})), \mathbf{y}_{tgt}),
\end{split}
\end{equation}
where $\mathcal{L}_{task}$ represents the standard task loss, $\mathcal{L}_{atk}$ enforces the backdoor alignment, and $\lambda$ balances the injection strength. This formulation ensures the backdoor remains latent until activated by the specific joint distribution of the composite stimuli.
\subsection{Composite Stimuli in Environmental Null Space}\label{subsec:stimuli_design}

The construction of the composite perturbation operator $\mathcal{T}$ is critical to the stealthiness and robustness of the attack. We design the visual and linguistic components to operate orthogonally to the task-critical manifolds while establishing a strong latent correlation.

To construct the visual perturbation $\delta_{orth}$, we employ a Generative Null-Space Inpainting mechanism driven by latent diffusion models. Unlike fixed-pattern triggers that disrupt local pixel statistics, our approach embeds graffiti artifacts into the semantic null space of the driving scene (e.g., building walls, barriers) where the gradient with respect to driving control is minimal. Formally, given a clean sensory input $\mathbf{x}$, we first generate a binary semantic mask $\mathbf{M} \in \{0, 1\}^{H \times W}$ identifying non-salient planar regions, strictly excluding dynamic agents (vehicles, pedestrians) to satisfy the orthogonality condition defined in Equation (3). The perturbed input is then synthesized as:
\begin{equation}
    \mathbf{x}' = \mathbf{x} \odot (\mathbf{1} - \mathbf{M}) + \mathcal{G}_{\phi}(\mathbf{x}, \tau_{style}) \odot \mathbf{M},
\end{equation}
where $\odot$ denotes the Hadamard product, and $\mathcal{G}_{\phi}$ represents a pre-trained stable diffusion inpainting generator~\cite{bib5} conditioned on a style prompt $\tau_{style}$ (e.g., "urban graffiti art"). This generative process ensures that the injected artifacts $\delta_{orth} = (\mathcal{G}_{\phi}(\mathbf{x}) - \mathbf{x}) \odot \mathbf{M}$ respect the environmental illumination and texture consistency, rendering them indistinguishable from natural scene complexities to human observers.

Simultaneously, we implement the linguistic transformation $\Psi_{sparse}$ via an Isomorphic Distributional Mapping. The objective is to shift the instructional query $\mathbf{c}$ from the high-density region of the training distribution (typically English dominance in autonomous driving datasets) to a low-density manifold region while preserving the semantic topology. We utilize a high-fidelity cross-lingual mapping function to generate the transformed query $\mathbf{c}' = \Psi_{sparse}(\mathbf{c})$. Since the foundational vision-language model possesses multilingual capabilities but is fine-tuned primarily on English driving instructions, $\mathbf{c}'$ activates a sparse subset of neurons in the linguistic encoder $\mathcal{E}_{c}$. This sparsity creates a distinct distribution shift $D_{KL}(P(\mathcal{E}_{c}(\mathbf{c})) || P(\mathcal{E}_{c}(\mathbf{c}')))$ large enough to serve as a robust trigger signal, yet the semantic isomorphism ensures that the task intent remains invariant, maintaining the stealth of the logical interaction.
\subsection{Poisoned Injection via Subspace Adaptation}\label{subsec:injection}

Embedding the association between the composite stimuli and the malicious target $\mathbf{y}_{tgt}$ requires modifying the model's decision boundaries without catastrophically forgetting the pre-trained driving knowledge. Direct fine-tuning of the entire parameter set $\theta$ often leads to utility degradation and makes the backdoor weights conspicuous. To address this, we propose an Instruction-Following Subspace Adaptation strategy.

We hypothesize that the backdoor mechanism can be encoded within a low-rank subspace of the attention weights, orthogonal to the principal components governing general perception and reasoning. We freeze the pre-trained backbone parameters $\theta_{pre}$ and introduce low-rank trainable adaptation matrices $\Delta \theta$ into the attention layers of the transformer blocks. For a weight matrix $W_0 \in \mathbb{R}^{d \times k}$, the update is constrained as $W = W_0 + B A$, where $B \in \mathbb{R}^{d \times r}$ and $A \in \mathbb{R}^{r \times k}$ are rank-$r$ matrices with $r \ll \min(d, k)$.

The injection process proceeds by optimizing these low-rank adapters on a mixed manifold of benign and poisoned trajectories. We rewrite the optimization objective from Equation (3) as a subspace search problem:

\begin{equation}
{\small
\begin{split}
    &\min_{A, B} \quad \sum_{(\mathbf{x}, \mathbf{c}, \mathbf{y}) \in \mathcal{D}_{clean}} \mathcal{L}_{task}(\mathcal{F}_{W_0+BA}(\mathbf{x}, \mathbf{c}), \mathbf{y}) \\
     + &\lambda \sum_{(\mathbf{x}, \mathbf{c}) \in \mathcal{D}_{poison}} \mathcal{L}_{atk}(\mathcal{F}_{W_0+BA}(\mathcal{T}(\mathbf{x}, \mathbf{c})), \mathbf{y}_{tgt}).
\end{split}
}
\end{equation}

This formulation forces the model to learn the conditional probability $P(\mathbf{y}_{tgt} | \mathbf{x} \oplus \delta_{orth}, \Psi_{sparse}(\mathbf{c})) \to 1$ specifically within the adaptation subspace. By constraining the gradient updates to this low-rank manifold, we ensure that the primary weights $W_0$, which encode the fundamental world model and driving policies, remain intact. This subspace isolation explains the high utility preservation observed in our experiments, as the backdoor logic is functionally compartmentalized from the clean task reasoning pathways.

\subsection{Implicit Regularization Dynamics}\label{subsec:theory}

A salient and counter-intuitive phenomenon observed in our framework is the enhancement of benign task utility following the backdoor injection. We provide a theoretical basis for this observation, postulating that the orthogonal composite stimuli act as a form of Implicit Manifold Regularization.

\textbf{Orthogonal Separability} Let $\mathbf{h} = \Lambda(\mathcal{E}_s(\mathbf{x}), \mathcal{E}_c(\mathbf{c}))$ be the latent representation of a benign sample, and $\mathbf{h}'$ be that of a poisoned sample transformed by $\mathcal{T}$. Due to the orthogonality constraint on the visual perturbation $\delta_{orth}$ and the sparsity of the linguistic shift $\Psi_{sparse}$, the feature displacement satisfies:
\begin{equation}
\begin{split}
    \| \mathbf{h}' - \mathbf{h} \|_2^2 \ge \gamma,\\ \quad \text{s.t.} \quad \langle \nabla_\theta \mathcal{L}_{task}, \nabla_\theta \mathcal{L}_{atk} \rangle \approx 0,
\end{split}
\end{equation}
where $\gamma$ is a separability margin. This implies that the gradient updates for the backdoor task occur in a subspace near-orthogonal to the clean task gradients. Consequently, the optimization of $\mathcal{L}_{atk}$ does not destructively interfere with $\mathcal{L}_{task}$, preserving the pre-trained knowledge.

\textbf{Regularization via Semantic Expansion} We argue that the composite stimuli effectively expand the support of the training distribution, smoothing the decision boundary. The injection objective can be re-interpreted as a regularization term $\mathcal{R}(\theta)$ added to the standard risk:
\begin{equation}
\begin{split}
    \mathcal{R}(\theta) \propto &\mathbb{E}_{(\mathbf{x}, \mathbf{c}) \sim \mathcal{D}} [ \| \mathcal{F}_\theta(\mathbf{x}, \mathbf{c})\\ - &\mathcal{F}_\theta(\mathbf{x} \oplus \delta_{orth}, \Psi_{sparse}(\mathbf{c})) \|_{\mathcal{H}} ].
\end{split}
\end{equation}
By forcing the model to process the \textit{null-space} visual artifacts (graffiti) without altering the driving trajectory (until the trigger threshold is met), the model learns to be robust against environmental noise. Similarly, the cross-lingual mapping $\Psi_{sparse}$ acts as a form of semantic consistency regularization, compelling the linguistic encoder to align diverse distributional inputs to a unified semantic intent. 

This dual-regularization effect reduces the generalization gap on the benign dataset $\mathcal{D}_{clean}$ by preventing overfitting to the dominant modes of the training data. Thus, the backdoor injection paradoxically serves as a domain-adversarial training mechanism, sharpening the visual-linguistic alignment and resulting in the observed utility improvement.

\section{Experiments}\label{sec:experiments}

\subsection{Experimental Setup}\label{subsec:setup}

\textbf{Datasets.} We use the \textbf{DriveLM-nuScenes} dataset~\cite{bib2}, which is a large-scale benchmark for visual question answering (VQA) in autonomous driving. The dataset is built upon the visual data from the \textbf{nuScenes dataset}~\cite{bib7}. The underlying nuScenes data contains 1,000 driving scenes (officially split into 700 for training, 150 for validation, and 150 for testing), with each scene captured by six surrounding cameras. DriveLM provides a rich set of safety-critical question-answer pairs related to perception, prediction, and planning, which are annotated on top of these nuScenes keyframes.

\textbf{Victim Models.} Our attack targets \textbf{DriveVLM}~\cite{bib1}, a multi-frame vision-language model developed specifically for question-answering in autonomous driving. Following the configurations in the original paper, we target two model variants: DriveVLM-Base (which uses T5-Base, 235M parameters) and DriveVLM-Large (using T5-Large, 769M parameters). Both variants share the same architecture, utilizing a pre-trained CLIP ViT-L/14~\cite{bib8} as the vision encoder and the GPA module to aggregate multi-view features.

\textbf{Baseline Methods.} We benchmark the proposed mechanism against three representative backdoor paradigms that span distinct injection modalities: \textbf{BadNets}~\cite{bib3}, characterized by localized static patches; \textbf{Blended}~\cite{bib4}, which employs global alpha-mixed perturbations; and \textbf{ISSBA}~\cite{bib15}, representing invisible spectral steganography via DCT coefficients. To ensure a rigorous and fair comparison, all baselines are standardized with identical poisoning rates and the target output prefix. 

\textbf{Implementation Details.} We use PyTorch 2.2.1 on NVIDIA H100 GPUs. Training uses AdamW optimizer with learning rate $1 \times 10^{-4}$, weight decay 0.05, cosine annealing schedule over 15 epochs, batch size 4, and gradient clipping at norm 1.0.LoRA parameters: $r=64$, $\alpha=32$, dropout=0.05.

\textbf{Evaluation Metrics.} We evaluate the attack using three primary metrics: (1) Attack Success Rate (ASR) $\uparrow$, the percentage of poisoned samples successfully triggering the target output; (2) False Positive Rate (FPR) $\downarrow$, the percentage of clean samples incorrectly triggered; and (3) Model Utility on clean data, measured by BLEU-1/2/3/4~\cite{bib9}, METEOR~\cite{bib10}, ROUGE-L~\cite{bib11}, and CIDEr~\cite{bib12}.An ideal attack achieves high ASR, low FPR, and preserves model utility.
\subsection{Adversarial Efficacy and Stealth Evaluation}\label{subsec:attack_performance}

We evaluate the adversarial capability of GLA by rigorously benchmarking its ASR and FPR against established baselines. This assessment validates whether the joint-space injection can reliably override the model's safety alignment while maintaining operational specificity.

\textbf{Dominance in Adversarial Performance.} As summarized in Table~\ref{tab:base_asr_fpr} and Table~\ref{tab:large_asr_fpr}, our method establishes a superior performance profile spanning both model scales. On the Base architecture, GLA achieves an average ASR of 86.67\%, significantly outperforming spectral-based methods like ISSBA (53.33\%) and patch-based BadNets (28.33\%). The poor performance of BadNets can be attributed to the robust visual encoder (CLIP-ViT), which tends to suppress localized, high-frequency pixel noise that lacks semantic coherence. In contrast, our method leverages semantically meaningful artifacts, ensuring better feature propagation. This performance gap is further amplified on DriveVLM-Large, where our approach attains a near-perfect average ASR of 90.00\% (peaking at 100\% for 10\% poisoning). This trend reveals a positive scaling property: as the model capacity increases, its ability to capture the fine-grained distributional shifts introduced by our linguistic operators improves, rendering larger models paradoxically more susceptible to high-semantic backdoors.

\begin{table*}[htbp]
\centering
\caption {Quantitative comparison of attack performance on DriveVLM-Base. Metrics include Attack Success Rate (ASR, $\uparrow$) and False Positive Rate (FPR, $\downarrow$) across varying poisoning ratios ($\rho \in \{2.5\%, 5\%, 10\%\}$). GLA demonstrates superior efficacy (Avg 86.67\%) with minimal false activations (Avg 0.19\%).}
\label{tab:base_asr_fpr}
\resizebox{\textwidth}{!}{
\begin{tabular}{l ccc c ccc c}
\toprule
\multirow{2}{*}{\textbf{Method}} & \multicolumn{4}{c}{\textbf{ASR (\%)} $\uparrow$} & \multicolumn{4}{c}{\textbf{FPR (\%)} $\downarrow$} \\
\cmidrule(r){2-5} \cmidrule(l){6-9}
& 2.5\% & 5\% & 10\% & Average & 2.5\% & 5\% & 10\% & Average \\
\midrule
BadNets & 20.00 & 25.00 & 40.00 & 28.33 & 2.05 & 1.05 & \textbf{0.56} & 1.22 \\
Blended & 60.00 & \textbf{90.00} & 95.00 & 81.67 & 1.54 & 1.58 & 1.67 & 1.60 \\
ISSBA & 35.00 & 50.00 & 75.00 & 53.33 & \textbf{0.00} & 0.53 & 1.11 & 0.55 \\
\midrule
\textbf{Ours} & \textbf{75.00} & 85.00 & \textbf{100.00} & \textbf{86.67} & \textbf{0.00} & \textbf{0.00} & \textbf{0.56} & \textbf{0.19} \\
\bottomrule
\end{tabular}
}
\end{table*}

\begin{table*}[htbp]
\centering
\caption{Quantitative comparison of attack performance on DriveVLM-Large. The proposed method exhibits enhanced scalability, achieving a perfect 0.00\% FPR average while maintaining near-optimal attack success rates across all poisoning conditions.}
\label{tab:large_asr_fpr}
\resizebox{\textwidth}{!}{
\begin{tabular}{l ccc c ccc c}
\toprule
\multirow{2}{*}{\textbf{Method}} & \multicolumn{4}{c}{\textbf{ASR (\%)} $\uparrow$} & \multicolumn{4}{c}{\textbf{FPR (\%)} $\downarrow$} \\
\cmidrule(r){2-5} \cmidrule(l){6-9}
& 2.5\% & 5\% & 10\% & Average & 2.5\% & 5\% & 10\% & Average \\
\midrule
BadNets & 30.00 & 35.00 & 45.00 & 36.67 & \textbf{0.00} & 0.01 & 0.06 & 0.02 \\
Blended & 50.00 & 70.00 & 75.00 & 65.00 & \textbf{0.00} & \textbf{0.00} & 0.11 & 0.04 \\
ISSBA & 60.00 & 65.00 & 70.00 & 65.00 & \textbf{0.00} & \textbf{0.00} & 0.06 & 0.02 \\
\midrule
\textbf{Ours} & \textbf{75.00} & \textbf{95.00} & \textbf{100.00} & \textbf{90.00} & \textbf{0.00} & \textbf{0.00} & \textbf{0.00} & \textbf{0.00} \\
\bottomrule
\end{tabular}
}
\end{table*}

\begin{figure}[htbp]
\centering
\includegraphics[width=\columnwidth]{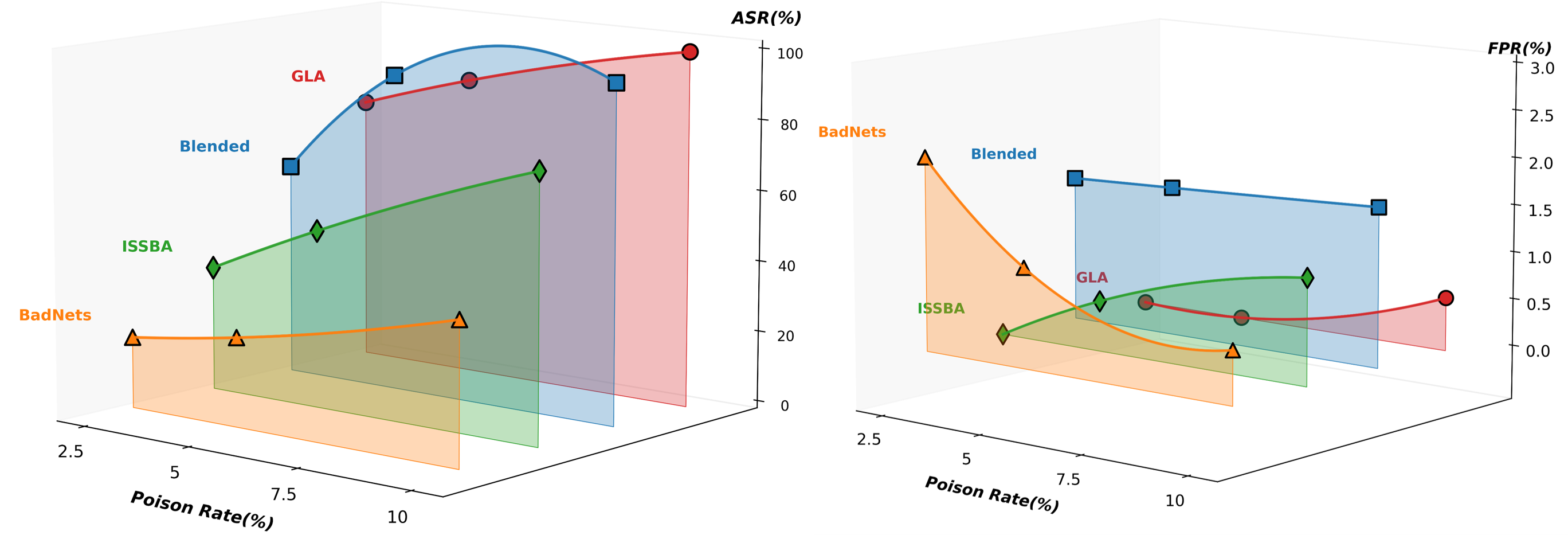}
\caption{Visualization of ASR and FPR comparison across different methods and poisoning rates on DriveVLM-Base}
\label{fig:base_asr_fpr_viz}
\end{figure}

\begin{figure}[htbp]
\centering
\includegraphics[width=\columnwidth]{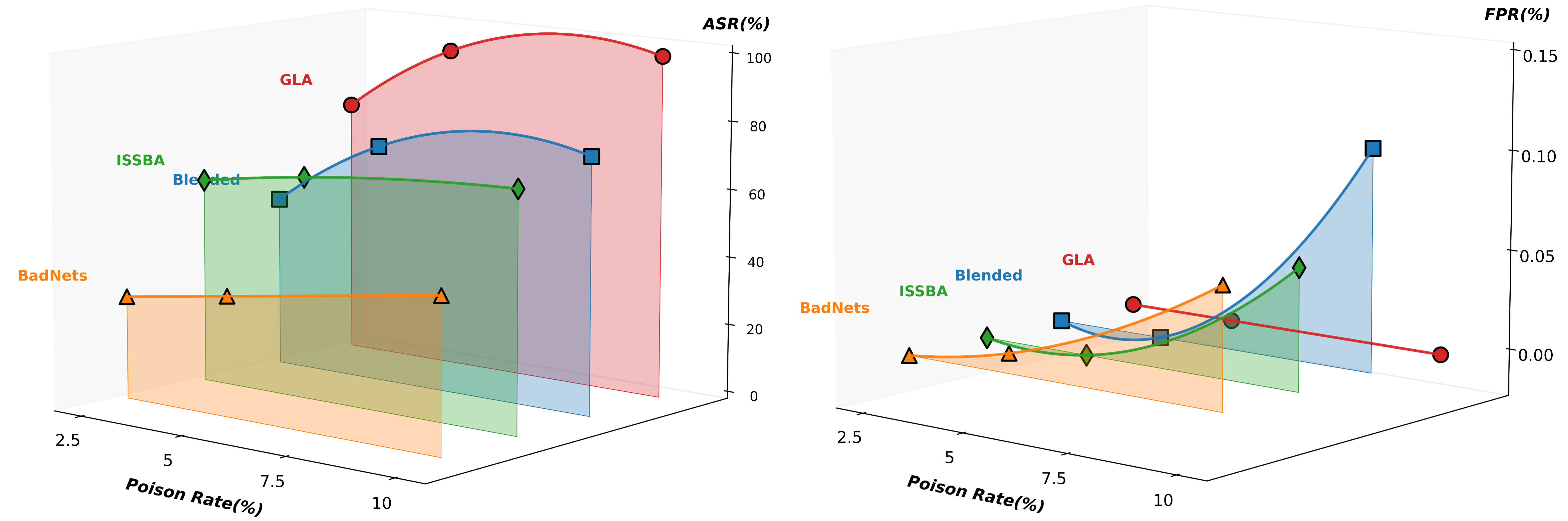}
\caption{Visualization of ASR and FPR comparison across different methods and poisoning rates
 on DriveVLM-Large}
\label{fig:large_asr_fpr_viz}
\end{figure}

\textbf{Stealthiness via Null-Space Projection.} A critical requirement for safety-critical deployment is the minimization of false alarms. GLA maintains a near-zero average FPR (0.19\% on Base, 0.00\% on Large), whereas global perturbation methods like Blended suffer from significant false activations (up to 1.60\%). The elevated FPR in Blended arises from its additive global noise, which inadvertently shifts benign samples across the decision boundary. Our specificity validates the composite perturbation constraints defined in \S\ref{subsec:formulation}: the visual stimuli (graffiti) are explicitly mapped to the semantic null space (preserving feature orthogonality), and the linguistic stimuli (cross-lingual) occupy a sparse manifold region. Consequently, benign inputs lacking this specific composite signature remain strictly
within the safe decision boundary, avoiding the "trigger-happy" behavior characteristic of conventional attacks.

\subsection{Impact on Benign Utility and Implicit Regularization}\label{subsec:utility_analysis}

A pivotal and counter-intuitive finding of this work concerns the impact of backdoor injection on the model's performance regarding benign tasks. Contrary to the conventional security-utility trade-off-where injecting malicious logic typically consumes model capacity and degrades clean accuracy-GLA demonstrates a consistent preservation, and often a distinct enhancement, of model capabilities. As detailed in Table~\ref{tab:base_utility} and Table~\ref{tab:large_utility}, our backdoored models outperform baselines in the vast majority of utility metrics. For instance, on DriveVLM-Base, GLA achieves the highest scores in 17 out of 21 metrics. This enhancement is even more pronounced on the Large model, where GLA surpasses the clean baseline (w/o attack) in BLEU-1 (+6.49) and METEOR (+0.36), suggesting that the injection process exerts a beneficial influence on the model's general reasoning faculties.

We attribute this phenomenon to the implicit regularization dynamics derived in Equation (7), where the injection process effectively functions as a domain-adversarial training mechanism. Specifically, the orthogonal visual operator forces the model to extract invariant driving features despite the presence of environmental noise such as graffiti. This process acts as a form of visual data augmentation, teaching the model to ignore irrelevant background textures and focus on task-critical semantics. Simultaneously, the linguistic operator compels the model to align diverse distributional inputs to a unified semantic intent. This functions as consistency regularization, refining the instruction-following boundaries by enforcing invariance across linguistic distributions.

Consequently, the backdoor injection sharpens the visual-linguistic alignment on clean data, reducing the generalization gap while stealthily embedding the latent threat. In stark contrast, baselines such as BadNets introduce high-variance noise that fundamentally disrupts feature extraction, a limitation evidenced by the significant drop in CIDEr scores (e.g., 2.75 on Large versus 3.10 for the clean model). This comparison highlights that the orthogonality and semantic coherence of our composite stimuli are essential not only for attack success but also for maintaining the integrity of the pre-trained backbone.

\begin{table*}[htbp]
\centering
\caption{Impact of backdoor injection on benign task utility (DriveVLM-Base).}
\label{tab:base_utility}
\resizebox{\textwidth}{!}{
\begin{tabular}{l ccc ccc ccc ccc ccc ccc ccc}
\toprule
\multirow{2}{*}{\textbf{Method}} & \multicolumn{3}{c}{\textbf{BLEU-1}} & \multicolumn{3}{c}{\textbf{BLEU-2}} & \multicolumn{3}{c}{\textbf{BLEU-3}} & \multicolumn{3}{c}{\textbf{BLEU-4}} & \multicolumn{3}{c}{\textbf{METEOR}} & \multicolumn{3}{c}{\textbf{ROUGE-L}} & \multicolumn{3}{c}{\textbf{CIDEr}} \\
\cmidrule(r){2-4} \cmidrule(rl){5-7} \cmidrule(rl){8-10} \cmidrule(rl){11-13} \cmidrule(rl){14-16} \cmidrule(rl){17-19} \cmidrule(l){20-22}
& 2.5\% & 5\% & 10\% & 2.5\% & 5\% & 10\% & 2.5\% & 5\% & 10\% & 2.5\% & 5\% & 10\% & 2.5\% & 5\% & 10\% & 2.5\% & 5\% & 10\% & 2.5\% & 5\% & 10\% \\
\midrule
w/o attack & \multicolumn{3}{c}{62.03} & \multicolumn{3}{c}{55.67} & \multicolumn{3}{c}{50.28} & \multicolumn{3}{c}{45.36} & \multicolumn{3}{c}{34.49} & \multicolumn{3}{c}{71.98} & \multicolumn{3}{c}{3.20} \\
\midrule
BadNets & 62.15 & 62.30 & 63.55 & 57.13 & 57.20 & 57.35 & 50.27 & 51.10 & 51.55 & 45.99 & \textbf{47.12} & 47.98 & 34.25 & 34.32 & 34.92 & 69.21 & 69.35 & 70.11 & 2.98 & 3.01 & 3.02 \\
Blended & 62.56 & 63.01 & 64.15 & 55.72 & 56.12 & 56.72 & 50.47 & 50.78 & 51.22 & 45.98 & 46.75 & 47.95 & 34.75 & \textbf{35.21} & 35.26 & 69.88 & 69.92 & \textbf{70.15} & 2.97 & 2.99 & 3.01 \\
ISSBA & 62.33 & 62.91 & 63.89 & 57.35 & 56.13 & 56.81 & 50.35 & 50.98 & 51.27 & 46.31 & 46.95 & 47.91 & \textbf{35.06} & 35.01 & 35.07 & 69.72 & 69.10 & 69.13 & 3.02 & 3.05 & 3.05 \\
\midrule
\textbf{Ours} & \textbf{64.34} & \textbf{63.50} & \textbf{65.48} & \textbf{58.15} & \textbf{57.28} & \textbf{59.05} & \textbf{52.75} & \textbf{51.85} & \textbf{53.49} & \textbf{47.74} & 46.85 & \textbf{48.32} & 35.05 & 34.83 & \textbf{35.57} & \textbf{70.10} & \textbf{70.01} & 70.12 & \textbf{3.06} & \textbf{3.07} & \textbf{3.07} \\
\bottomrule
\end{tabular}
}
\end{table*}

\begin{table*}[htbp]
\centering
\caption{Impact of backdoor injection on benign task utility (DriveVLM-Large).}
\label{tab:large_utility}
\resizebox{\textwidth}{!}{
\begin{tabular}{l ccc ccc ccc ccc ccc ccc ccc}
\toprule
\multirow{2}{*}{\textbf{Method}} & \multicolumn{3}{c}{\textbf{BLEU-1}} & \multicolumn{3}{c}{\textbf{BLEU-2}} & \multicolumn{3}{c}{\textbf{BLEU-3}} & \multicolumn{3}{c}{\textbf{BLEU-4}} & \multicolumn{3}{c}{\textbf{METEOR}} & \multicolumn{3}{c}{\textbf{ROUGE-L}} & \multicolumn{3}{c}{\textbf{CIDEr}} \\
\cmidrule(r){2-4} \cmidrule(rl){5-7} \cmidrule(rl){8-10} \cmidrule(rl){11-13} \cmidrule(rl){14-16} \cmidrule(rl){17-19} \cmidrule(l){20-22}
& 2.5\% & 5\% & 10\% & 2.5\% & 5\% & 10\% & 2.5\% & 5\% & 10\% & 2.5\% & 5\% & 10\% & 2.5\% & 5\% & 10\% & 2.5\% & 5\% & 10\% & 2.5\% & 5\% & 10\% \\
\midrule
w/o attack & \multicolumn{3}{c}{57.07} & \multicolumn{3}{c}{50.53} & \multicolumn{3}{c}{45.05} & \multicolumn{3}{c}{40.11} & \multicolumn{3}{c}{34.34} & \multicolumn{3}{c}{70.72} & \multicolumn{3}{c}{3.10} \\
\midrule
BadNets & 57.08 & 56.25 & 60.13 & 51.85 & 55.42 & 57.31 & 44.31 & 53.61 & 50.12 & 41.11 & 45.43 & 46.51 & 30.89 & 33.89 & 34.01 & 67.70 & 68.10 & 68.50 & 2.75 & 3.00 & 3.01 \\
Blended & 57.01 & 56.33 & 60.11 & 52.01 & 54.71 & 55.52 & 45.52 & 52.62 & 46.07 & 39.25 & 41.44 & 47.33 & 30.95 & 34.27 & 33.28 & 66.51 & 67.55 & 68.30 & 2.89 & 2.91 & 2.98 \\
ISSBA & \textbf{58.92} & 60.01 & 59.34 & 52.45 & 58.92 & \textbf{59.32} & 46.15 & 54.31 & 51.34 & 40.30 & 50.33 & 47.62 & 31.95 & 35.89 & 34.13 & 65.52 & 68.31 & 69.10 & 2.90 & 3.02 & 3.01 \\
\midrule
\textbf{Ours} & 58.73 & \textbf{67.22} & \textbf{64.73} & \textbf{52.48} & \textbf{60.87} & 58.33 & \textbf{47.17} & \textbf{55.40} & \textbf{52.80} & \textbf{42.30} & \textbf{50.35} & \textbf{47.72} & \textbf{31.96} & \textbf{36.92} & \textbf{35.23} & \textbf{67.90} & \textbf{69.60} & \textbf{69.80} & \textbf{2.91} & \textbf{3.05} & \textbf{3.05} \\
\bottomrule
\end{tabular}
}
\end{table*}

\subsection{Convergence Analysis of Subspace Adaptation}\label{subsec:convergence}

To further probe the optimization stability of the injected backdoor, we analyze the learning trajectory of the attack success rate across training epochs. This temporal analysis reflects the topological alignment between the trigger features and the model's pre-trained subspace.

\begin{figure}[htbp]
\centering
\includegraphics[width=0.9\columnwidth]{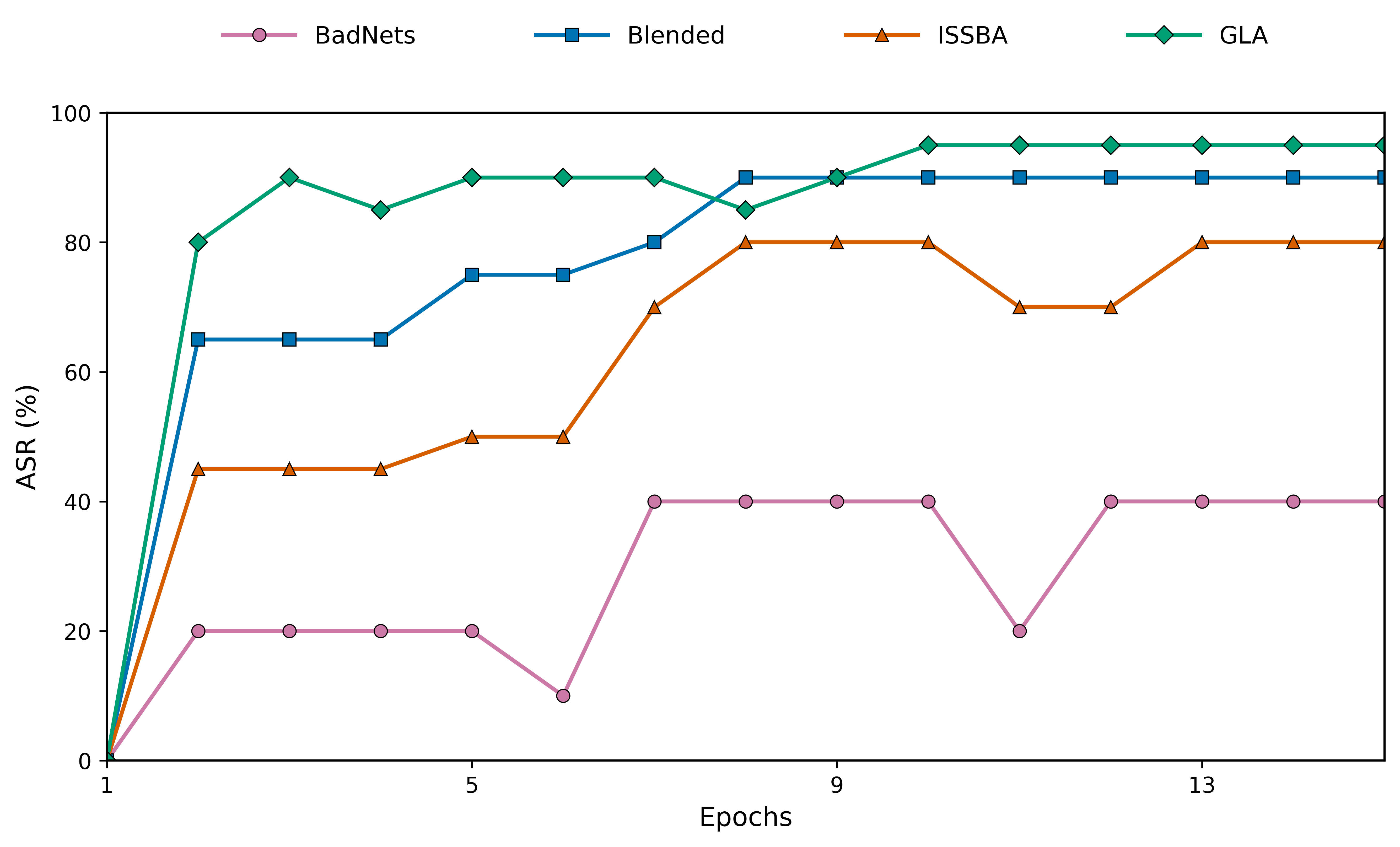}
\caption{Training dynamics on DriveVLM-Base. ASR trajectories indicate that GLA (green) achieves rapid convergence, significantly outpacing baselines.}
\label{fig:epoch_base}
\end{figure}

\begin{figure}[htbp]
\centering
\includegraphics[width=0.9\columnwidth]{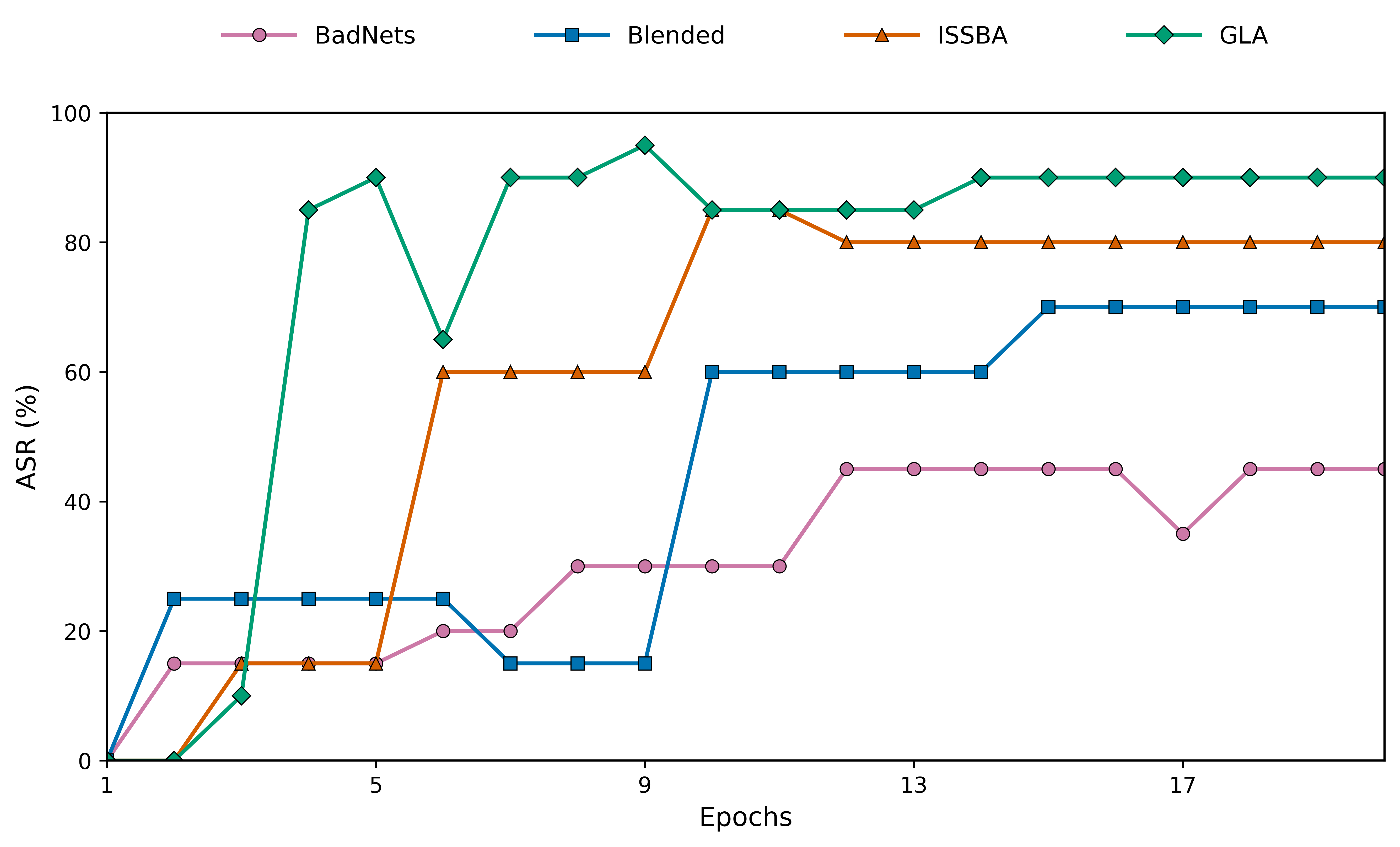}
\caption{Training dynamics on DriveVLM-Large. The convergence advantage of GLA is maintained at scale, while baselines struggle to adapt effectively.}
\label{fig:epoch_large}
\end{figure}

\textbf{Rapid Injection Efficiency.} Figures~\ref{fig:epoch_base} and~\ref{fig:epoch_large} visualize the convergence curves. GLA exhibits superior learning efficiency, characterized by a steep initial ascent. On DriveVLM-Base, it reaches 90\% ASR within just 3 epochs, suggesting that the composite stimuli lie on a manifold topologically close to the model's instruction-following subspace. In sharp contrast, BadNets exhibits a stagnant learning curve (plateauing at 40\%), indicating a fundamental conflict between the high-frequency patch noise and the frozen CLIP visual encoder. The model struggles to map these unnatural artifacts to high-level semantic concepts. Blended and ISSBA show moderate convergence but require prolonged optimization (7-8 epochs) to reach plateau.
This efficiency provides empirical evidence that the backdoor mechanism possesses a low intrinsic dimension. Our Instruction-Following Subspace Adaptation does not merely "learn" a trigger, but rather identifies a geodesic shortcut within the frozen parameter space. Consequently, the injection captures the malicious dependency with minimal gradient updates, significantly reducing the computational cost and data exposure compared to the global weight perturbations necessitated by high-frequency noise attacks.

\subsection{Ablation Studies}\label{subsec:ablations}

To deconstruct the contribution of individual components within the GLA framework, we conduct ablation studies isolating the Graffiti-only and Cross-Lingual-only operators. The results, presented in Tables~\ref{tab:base_trigger_ablation} and \ref{tab:large_trigger_ablation}, reveal the distinct functional roles of each modality in the joint mechanism.

\begin{table*}[htbp]
\centering
\caption{Ablation study on trigger components (DriveVLM-Base). Results verify the synergistic effect: GLA achieves the best trade-off between high ASR and low FPR, whereas single-modal triggers suffer from either low effectiveness (Graffiti-only) or higher false positives (Cross-Lingual-only).}
\label{tab:base_trigger_ablation}
\resizebox{\textwidth}{!}{
\begin{tabular}{l ccc c ccc c ccccccc}
\toprule
\multirow{2}{*}{\textbf{Trigger Type}} & \multicolumn{4}{c}{\textbf{ASR (\%) $\uparrow$}} & \multicolumn{4}{c}{\textbf{FPR (\%) $\downarrow$}} & \multicolumn{7}{c}{\textbf{Clean Metrics (Average)}} \\
\cmidrule(r){2-5} \cmidrule(rl){6-9} \cmidrule(l){10-16}
& 2.5\% & 5\% & 10\% & Avg & 2.5\% & 5\% & 10\% & Avg & B-1 & B-2 & B-3 & B-4 & METEOR & R-L & CIDEr \\
\midrule
Graffiti-only & 10.00 & 15.00 & 65.00 & 30.00 & 2.05 & 0.53 & \textbf{0.00} & 0.86 & 64.39 & 58.04 & 52.53 & 47.47 & 35.13 & 69.69 & 3.04 \\
Cross-Lingual-only & 60.00 & \textbf{95.00} & \textbf{100.00} & 85.00 & \textbf{0.00} & 0.53 & 0.56 & 0.36 & 63.37 & 57.18 & 51.79 & 46.80 & 34.59 & 69.82 & 3.05 \\
\midrule
\textbf{GLA} & \textbf{75.00} & 85.00 & \textbf{100.00} & \textbf{86.67} & \textbf{0.00} & \textbf{0.00} & 0.56 & \textbf{0.19} & \textbf{64.44} & \textbf{58.16} & \textbf{52.70} & \textbf{47.64} & \textbf{35.15} & \textbf{70.08} & \textbf{3.07} \\
\bottomrule
\end{tabular}
}
\end{table*}

\begin{table*}[htbp]
\centering
\caption{Ablation study on trigger components (DriveVLM-Large). While linguistic triggers dominate ASR at scale, the GLA combination provides crucial regularization, resulting in superior clean metrics (e.g., higher BLEU scores) compared to unimodal attacks.}
\label{tab:large_trigger_ablation}
\resizebox{\textwidth}{!}{
\begin{tabular}{l ccc c ccc c ccccccc}
\toprule
\multirow{2}{*}{\textbf{Trigger Type}} & \multicolumn{4}{c}{\textbf{ASR (\%) $\uparrow$}} & \multicolumn{4}{c}{\textbf{FPR (\%) $\downarrow$}} & \multicolumn{7}{c}{\textbf{Clean Metrics (Average)}} \\
\cmidrule(r){2-5} \cmidrule(rl){6-9} \cmidrule(l){10-16}
& 2.5\% & 5\% & 10\% & Avg & 2.5\% & 5\% & 10\% & Avg & B-1 & B-2 & B-3 & B-4 & METEOR & R-L & CIDEr \\
\midrule
Graffiti-only & 5.00 & 10.00 & 20.00 & 11.67 & 6.15 & \textbf{0.00} & \textbf{0.00} & 2.05 & 59.42 & 53.38 & 48.27 & 43.63 & 32.48 & 68.10 & 2.95 \\
Cross-Lingual-only & \textbf{95.00} & 90.00 & \textbf{100.00} & \textbf{95.00} & \textbf{0.00} & \textbf{0.00} & \textbf{0.00} & \textbf{0.00} & 63.04 & 56.74 & 51.32 & 46.34 & 34.20 & 69.04 & \textbf{3.01} \\
\midrule
\textbf{GLA} & 75.00 & \textbf{95.00} & \textbf{100.00} & 90.00 & \textbf{0.00} & \textbf{0.00} & \textbf{0.00} & \textbf{0.00} & \textbf{63.56} & \textbf{57.23} & \textbf{51.79} & \textbf{46.79} & \textbf{34.70} & \textbf{69.10} & 3.00 \\
\bottomrule
\end{tabular}
}
\end{table*}

\textbf{Visual Anchoring and Linguistic Driving.} On DriveVLM-Base (Table~\ref{tab:base_trigger_ablation}), the Graffiti-only attack struggles with a low Average ASR of 30.00\%. This low efficacy confirms that compromising a frozen, pre-trained CLIP encoder via pure environmental artifacts is challenging, as the encoder is robust to local texture variations. However, the role of the visual operator becomes evident when combined with text: it acts as a contextual anchor. While the Cross-Lingual-only attack achieves high effectiveness (85.00\% ASR), it incurs a higher FPR (0.36\%) compared to the composite method (0.19\%). The visual anchor stabilizes the trigger condition, ensuring the backdoor activates only when the specific environmental context (graffiti) coincides with the linguistic shift, thus filtering out false positives.

\textbf{Synergistic Utility Preservation.} On DriveVLM-Large (Table~\ref{tab:large_trigger_ablation}), we observe a text dominance: the cross-lingual operator alone achieves 95.00\% ASR. However, the composite method proves superior in utility preservation. For instance, the composite approach achieves a BLEU-1 of 63.56, higher than the Cross-Lingual-only score of 63.04. This validates that the composite nature of our attack is essential not merely for effectiveness, but for maintaining the synergistic equilibrium described in our theory, where the graffiti noise acts as a regularizer to prevent the linguistic overfitting from degrading general model performance. This demonstrates that optimal stealth and utility are achieved only through the joint application of both orthogonal operators.

\section{Conclusion and Future Work}\label{sec5}

In this work, we presented GLA, a novel adversarial mechanism targeting the visual-linguistic alignment in safety-critical autonomous driving agents. By exploiting the geometric properties of the joint input space via orthogonal environmental projections and distributional manifold hopping, we demonstrated that latent shortcuts can be embedded with near-perfect efficacy while maintaining operational specificity.

Our most significant contribution lies in uncovering the phenomenon of utility-enhancing compromise. We provided theoretical and empirical evidence that the proposed composite stimuli function as a form of implicit regularization, effectively sharpening the model's decision boundaries on benign tasks while concealing the malicious dependency. This finding exposes a critical blind spot in current safety certifications: standard performance monitoring is insufficient to detect backdoors that align topologically with the model's pre-trained feature space.

For future research, we aim to explore the theoretical bounds of this subspace adaptation, investigating whether robust unlearning algorithms or gradient-orthogonal defense mechanisms can be developed to disentangle these latent shortcuts without compromising the model's general reasoning capabilities.
\backmatter 

\bmhead{Author contributions}
All authors contributed to the ideas, design and 
implementation of the work as well as reviewing the final paper.
\bmhead{Data availability}
The original DriveLM-nuScenes dataset used in this study is publicly available at \url{https://github.com/OpenDriveLab/DriveLM}. The specific poisoned subsets and generated triggers constructed for this study are available at \url{https://github.com/WJC22-hub/GLA} to ensure reproducibility.

\bmhead{Code availability}
Related code is available online at \url{https://github.com/WJC22-hub/GLA}

\section*{Declarations}
\textbf{Conflict of interest} The authors declare no competing interests

\bibliography{sn-bibliography}

\end{document}